\documentclass[runningheads]{llncs}
\usepackage{graphicx}
\usepackage[misc]{ifsym}
\usepackage{color}
\usepackage{ulem}
\usepackage{caption}
\usepackage{subfig}
\usepackage{amsfonts}
\usepackage{amsmath}
\usepackage{makecell}
\usepackage{multirow}
\usepackage{stfloats}
\usepackage{cite}
\usepackage{amsmath,amssymb,amsfonts}
\usepackage{algorithmic}
\usepackage{graphicx}
\usepackage{textcomp}
\usepackage{xcolor}
\begin{document}
\title{GLA-DA: Global-Local Alignment Domain Adaptation for Multivariate Time Series}
\titlerunning{GLA-DA for Multivariate Time Series}
%
\author{Gang Tu\inst{1} \and
Dan Li\Letter\inst{1}\and
Bingxin Lin\inst{1}\and
Zibin Zheng\inst{1} \and
See-Kiong Ng\inst{2,3}
\thanks{This paper has been accepted as a long paper in DASFAA2024.}
}
%
\authorrunning{G. Tu et al.}
%
\institute{
Sun Yat-Sen University, Zhuhai 519000, China\\
\email{\{tugang, linbx25\}@mail2.sysu.edu.cn, \{lidan263, zhzibin\}@mail.sysu.edu.cn}\\
\and
Institute of Data Science, National University of Singapore,
Singapore 117602, Singapore\\
\email{seekiong@nus.edu.sg}\\
\and
School of Computing, National University of Singapore,
Singapore 117417, Singapore
}

\maketitle              

\begin{abstract}

Unlike images and natural language tokens, time series data is highly semantically sparse, resulting in labor-intensive label annotations. Unsupervised and Semi-supervised Domain Adaptation (UDA and SSDA) have demonstrated efficiency in addressing this issue by utilizing pre-labeled source data to train on unlabeled or partially labeled target data. However, in domain adaptation methods designed for downstream classification tasks, directly adapting labeled source samples with unlabelled target samples often results in similar distributions across various classes, thereby compromising the performance of the target classification task. 
To tackle this challenge, we proposed a Global-Local Alignment Domain Adaptation (GLA-DA) method for multivariate time series data. Data from two domains were initially encoded to align in an intermediate feature space adversarially, achieving Global Feature Alignment (GFA). Subsequently, GLA-DA leveraged the consistency between similarity-based and deep learning-based models to assign pseudo labels to unlabeled target data. This process aims to preserve differences among data with distinct labels by aligning the samples with the same class labels together, achieving Local Class Alignment (LCA). We implemented GLA-DA in both UDA and SSDA scenarios, showcasing its superiority over state-of-the-art methods through extensive experiments on various public datasets. Ablation experiments underscored the significance of key components within GLA-DA.

\end{abstract}

\section{Introduction}
\label{sec: intro}
Time Series (TS) data can be recorded by various sensors and has been widely applied in many fields like healthcare monitoring, stock and weather forecasting, and manufacturing process control, etc. With the development and advance of Deep Learning (DL), researchers are increasingly leveraging its capabilities to uncover intricate dependencies within time series data \cite{Zhao: healthcare}\cite{He: forecasting}.  
Despite the influx of data facilitated by rapid advancements in sensing and storage technologies, a significant portion of time series data remains semantically sparse, predominantly unlabeled, and often requiring labor-intensive labeling processes. Standard supervised learning methods, which heavily depend on labeled data, exhibit inefficiencies under these circumstances. Additionally, these methods often operate under the assumption that training and test data share identical distributions. When deploying a pre-trained model in a new domain, the results can be sub-optimal due to the domain shift between the two domains.

To address the challenges posed by the high cost of manual annotation and the issue of domain shift, researchers have increasingly turned to Domain Adaptation (DA)—a prominent approach within transfer learning. This technique allows for the transfer of knowledge gleaned from a source domain to a target domain, facilitating downstream supervised learning tasks even in scenarios where labeled target data is insufficient. Existing domain adaptation methods include domain discrepancy minimization and domain discrimination. The former aims to reduce the distance, based on certain metrics, between the original or transformed data from two domains \cite{Weining: mmd}\cite{Zhang: djp-mmd}. In contrast, the latter seeks to extract domain-invariant features from both domains through adversarial means \cite{Lai: context-aware}.  
ErWe Tzeng et al. proposed the Adversarial Discriminative Domain Adaptation (ADDA) model to train a source encoder and classifier and then train a target encoder to align it with the embedded source data, where data from both domains confused the discriminator in an adversarial manner \cite{Eric: ADDA}. Han Zou et al. proposed the Consensus Adversarial Domain Adaptation(CADA) model to embed data from two domains into a new common feature space \cite{HAN: CADA}.

Many domain adaptation methods predominantly focus on visual tasks. Their backbones for extracting image features are LeNet \cite{lenet5}, AlexNet \cite{Alex: AlexNet} and Swin Transformer  \cite{Liu: SwinTransformer}, etc. These architectures significantly enhance the efficacy of algorithms applied to subsequent vision tasks, including domain adaptation. 
TS data inherently contains complex temporal dependencies that are DA tasks. Additionally, in multivariate TS data, spatial dependencies emerge, such as correlations among sensors and actuators deployed in smart building air-conditioning systems. These spatial-temporal dependencies could not be sufficiently captured by the aforementioned backbone models.
Furthermore, the majority of existing DA approaches pre-trained their models on ImageNet \cite{Ragab: SLARDA}, an extensive dataset comprising diverse images. In contrast, no equivalent comprehensive dataset exists specifically for TS data, making DA applications to TS data particularly challenging.

\begin{figure}
    \centering
    \subfloat[Class disappearing.]{\includegraphics[width=0.52\linewidth]{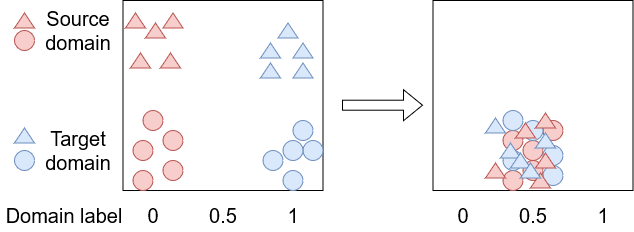}}
    \hspace{0.1in}
    \subfloat[Preserve classes differences.]{\includegraphics[width=0.43\linewidth]{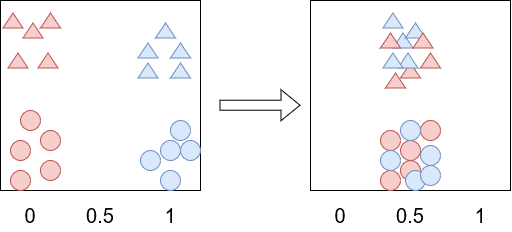}}
    \caption{Possible data distributions after global feature alignment.}
    
    \label{fig:distris}
\end{figure}

In addition, many current DA methods primarily focus on globally aligning the data distributions of two domains, overlooking the distinct distributional nuances among various classes. This oversight can lead to results after DA as shown in Fig.~\ref{fig:distris}-(a)), where distinct classes become intermixed after DA, thereby complicating the downstream classification tasks. Ideally, we aspire for post-DA samples from distinct classes to align cohesively as depicted in Fig.~\ref{fig:distris}-(b)). However, achieving class cohesiveness in the Unsupervised Domain Adaptation (UDA) scenario is challenging due to the absence of target labels. An ensemble teacher model was employed to produce target pseudo-labels in \cite{Ragab: SLARDA}. Only samples exceeding a certain prediction threshold were retained, thereby diminishing the beneficial impact of class-conditional shift when training the target classifier, which should have fully utilized target samples by assigning pseudo-labels.


Thus, a feasible DA method for Multivariate Time Series (MTS) data should tackle the following challenges: capturing temporal correlations for general multivariate time series data, maintaining maximum class cohesiveness, and fully leveraging the unlabelled target samples. To address these challenges, we proposed a novel Global-Local Alignment Domain Adaptation (GLA-DA) method for MTS data. GLA-DA aligns the data distributions of two domains by utilizing adversarial training to map them into an intermediate feature space, thereby achieving Global Feature Alignment (GFA). Subsequently, GLA-DA leverages the consistency between a similarity-based model and a deep learning-based model to assign pseudo labels to unlabeled target samples. This process aims to preserve differences among samples with distinct labels by aligning the label information, thus achieving Local Class Alignment (LCA). Additionally, we followed the settings of \cite{Ragab: ADATIME} and adopted 1D convolutions along the time axis to capture the temporal correlations of time series data. While time-series adaptation traditionally employs RNNs, researchers have empirically demonstrated the benefit of CNNs on sequential data \cite{Kiranyaz: 1dcnnsurvey}. We will show that our proposed method can benefit from a CNN architecture as explained in Section \ref{sec: exp}.

The main contributions of this work are as follows:

\begin{enumerate}
\item We proposed a novel unsupervised domain adaptation method to address the lack of labeled target samples via Global Feature Alignment (GFA) and Local Class Alignment (LCA). GFA was achieved by aligning data from two domains to an intermediate feature space and minimizing the adversarial loss through adversarial training. LCA was achieved by aligning samples with the same labels together and minimizing the class center loss.
\item We proposed a novel Agree Mechanism (AM) that leveraged the consistency between similarity-based and deep learning-based models to assign pseudo labels to unlabeled target data. AM was designed to generate as many target samples with rational pseudo labels to fully utilize target data resources. 
\item We conducted comprehensive experiments on various public time series datasets. Results showcase that the proposed GLA-DA  is superior to baselines. 


\end{enumerate}

The remaining parts of this paper are organized as follows. Section \ref{sec: rwork} summarizes related works. Section \ref{sec: method} states the procedure and key mathematical formulations. Section \ref{sec: exp} shows the setup and experimental results. Finally, section \ref{sec:conc} summarizes the work and proposes possible future work.

\section{Related Works}
\label{sec: rwork}
\subsection{Domain Adaptation} DA is well known for its ability to transfer the knowledge of the source domain model to the target domain by eliminating the influence of differences in data distribution and has drawn wide attention from researchers and has been applied mainly to vision tasks in recent years. Usually, DA is applied in unsupervised (unavailable to target labels) or semi-supervised (scarce target labels) scenarios. 

Some research realized DA using mathematical calculations to minimize the distribution differences based on some distance metric like Euclidean distance, cosine similarity, KL divergence, and maximum mean distance (MMD) \cite{Gretton: KTS}, etc. Some research also combines these metrics or creates a new one. For instance, \cite{Sun: MKSDA} simultaneously minimized MMD and the structural risk functional of support vector machines. \cite{Yin: MLADA} explored the relationship between a target sample's prediction probability and its distance to the decision boundary. \cite{Yan: newMMD} proposed weighted MMD and class-specific MMD to reduce the influence of class weight bias.

In recent years, researchers have been trying to utilize deep neural networks to obtain domain-invariant features. The common optimization goal is adversarial loss. It aligns data from two domains by deceiving a domain discriminator so that it cannot distinguish two domains. Some research aligns target data to the source domain while the other leads data or embedded features from two domains to a new feature space and our proposed GLA-DA implemented the latter. For example, \cite{Ganin: DANN} proposed a framework DANN that used a new gradient reversal layer to augment a feed-forward model and trained the neural network by competing with a domain regressor. \cite{Alanov: HDN} improved domain adaptation frameworks of GANs by introducing a compact parameter space for fine-tuning the generator. \cite{Su: AADA} aligned features from two domains in an adversarial manner and got sample importance weight by the prediction outputs of the domain discriminator for subsequent sample selection.

Other related research explores other approaches to extract domain invariant features or reconstruct target data. For example, \cite{Jian: UDATEBN} combined contrastive learning and uncorrelated conditioning for event-based data image classification. \cite{Huang: CaCo} constructed a semantics-aware dictionary by instance contrastive learning and learned domain invariant features in category contrast. \cite{Dhaini: DLDA} used dictionary learning to minimize the residue of reconstructed target data based on a learned source dictionary. However, their complexity leads to difficulty in generalization.

\subsection{DA on Time Series Data}
Due to the scarcity of works on domain adaptation for TS data, researchers have been exploring different approaches that can simultaneously transfer the knowledge from the source model and consider the temporal dependencies within TS data. \cite{Ragab: ADATIME} proposed a unified benchmarking suite ADATIME for DA on TS data, including backbones, datasets, and training schemes, and evaluated various CNN models on TS data to provide some backbones for reference. \cite{Ragab: SLARDA} introduced self-contrastive learning in the pre-training step \textcolor{black}{which can help the source encoder to be more sensitive to the source data} and aligned target data to the source data adversarially. \cite{Eldele: CoMix} proposed a mixup strategy that is robust to domain shift to generate two intermediate augmented views and make the distribution of each domain more similar to its corresponding augmented view to pull data from both domains to a common feature space. \textcolor{black}{Existing methods mainly explore adversarial training and constractive learning on DA on TS data to mitigate domain shift but make few efforts to implement local alignment after global alignment. Therefore, we proposed our GLA-DA to implement global alignment using adversarial learning and local alignment with pseudo labels.}


\section{Global-Local Alignment Domain Adaptation}
\label{sec: method}
\subsection{Architecture}
\begin{figure}[!t]
    \centering
    \includegraphics[width=1.0\linewidth]{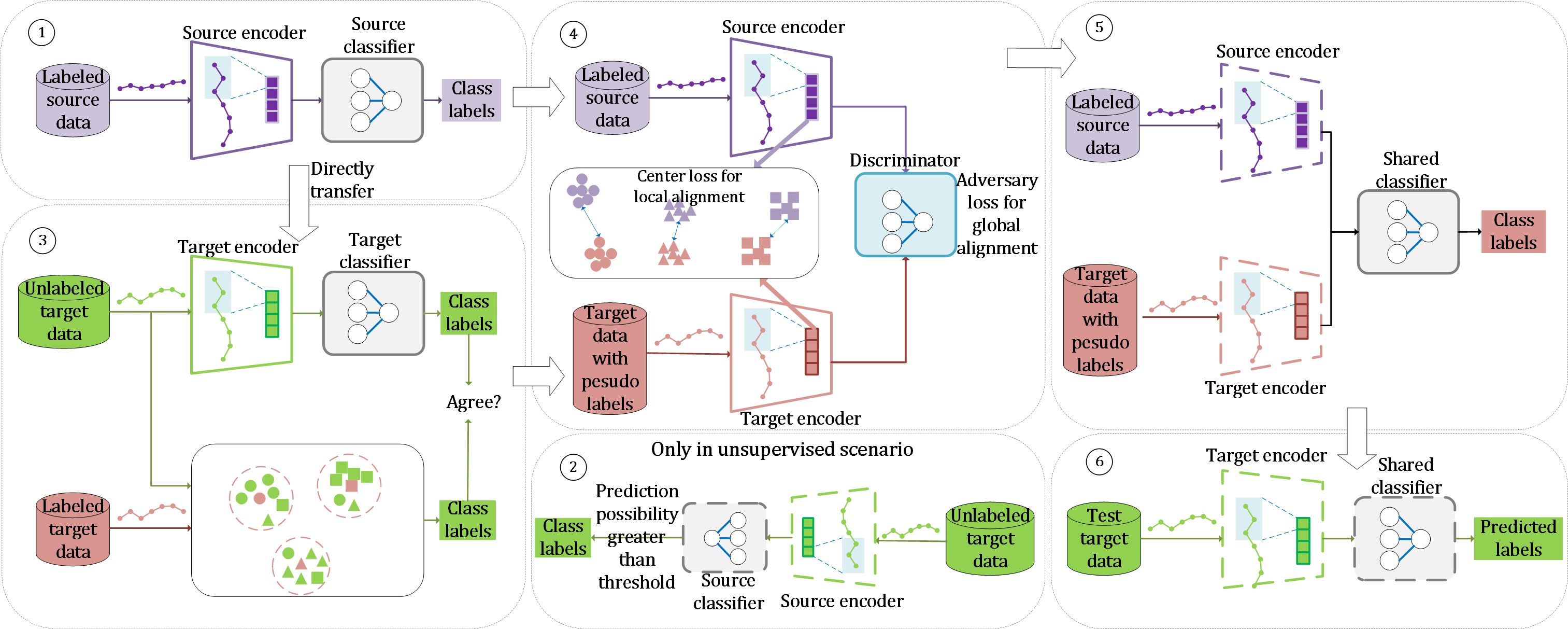}
    \caption{An overview of our model GLA-DA for Multivariate Time Series. The dashed lines indicate the parameters of that model are fixed. \textcolor{black}{First we trained a source model with labeled source data. Second, we assigned initial pseudo labels for unlabeled target dataset just in unsupervised scenario based on the threshold output by the source model. Third, we assigned pseudo labels to the unlabeled target samples left using AM. Forth, we performed global data alignment using adversarial learning and local alignment by optimizing center-loss. Last we trained a shared classifier.}}
    \label{fig:glada}
\end{figure}

Fig \ref{fig:glada} shows the schematic diagram of the proposed GLA-DA. First, the source encoder and source classifier are pre-trained based on the labeled source data. Then the target encoder is initialized with the parameters of its source counterpart due to the lack of labels in the UDA scenario. Subsequently, the logit layer of the source classifier (output probabilities) is applied to deal with the unlabelled target data. At this stage, only target samples whose probability values are bigger than a threshold will be assigned initial pseudo labels. Note that in the semi-supervised domain adaptation scenario, we will skip assigning initial pseudo labels 
\textcolor{black}{for there are initial labeled target samples offered, which meets the demand of assigning pseudo labels for the unlabeled target samples left}.
Then, upon the target samples with initial labels, a similarity-based clustering model and a deep neural network-based classification model deal with the remaining unlabeled target samples and assign labels, respectively. Here, the Agree Mechanism (AM) refers to the process of how we try to assign rational pseudo labels to as many unlabelled target samples as possible, which will be described with details in Section \ref{subsec:pseudo-labels}.
 Next, the labeled source data and the target data with sufficient pseudo-labels are adapted to an intermediate feature space. The discriminator is trained to distinguish encoded data from different domains, achieving Global Feature Alignment (GFA). Besides, the center loss is calculated to force data with the same labels to align together, thus achieving Local Class Alignment (LCA). This process will be described with details in Sections \ref{subsec:gfa} and \ref{subsec:lca}.
Finally, with globally and locally aligned source and target samples, a shared classifier is trained based on the adapted samples from both domains, which is subsequently utilized to classify target data.

In the following subsections, we will formulate the domain adaptation problem for multivariate time series data and drive the methodology with details. Details about implementations are provided in Sec.\ref{sec: exp}

\subsection{Problem Statement and Indications}
Given a tuple data set $(X, Y)$, where $X \in R^{p \times m \times n}$ is the time series data, where $p$ is the number of samples, $m$ is the dimension number and $n$ is the length of timestamp. and $Y \in R^{p \times 1}$ is the labels. The classifier could be trained by mapping $X^{tr}\in R^{k_{tr} \times m \times n}$ or its embedded features to the space of labeled data: $\mathcal{F}: X^{tr} \rightarrow Y^{tr}$. With enough training data in the same domain, the classification task could be accomplished by applying $\mathcal{F}$ to predict the corresponding labels of testing data $X^{te}$.

However, in practical cases, the testing data is usually collected from another domain where the trained source classifier $\mathcal{F}$ cannot be directly applied to the target data $X^{te}$. As such, we defined the source domain as $\mathcal{D}_s$ consists of feature space $\mathcal{X}_s \in R^{p_s \times m \times n}$ and a marginal probability $P(X_s)$, where $X_s=\{x_1,x_2,...,x_{p_{s}}\}$ is the set of input instances. Then we have $\mathcal{D}_s = \{\mathcal{X}_s,P(X_s)\}$. The source task $T_s$ uses a function $\mathcal{F}_s$ with $\mathcal{D}_s$ to predict labels of each instance in the label space $\mathcal{Y}_s$. For the target domain, we assume that each class of target data has the same small number of labeled samples. Then we could define $\mathcal{T}_{L} = \{X_{TL}, Y_{TL}\}$ as labeled target samples and $\mathcal{T}_{U} = \{X_{TU}, Y_{TU}\}$  as unlabeled target samples $|\mathcal{T}_{L}|$ was denoted as the number of labeled target samples and so as $|\mathcal{T}_{U}|$ for the unlabeled ones. 

We denote the source classification task as $T_s = \{\mathcal{Y}_s, \mathcal{F}_s(\cdot)\}$ with enough labelled source training data $\mathcal{D}_s = \{\mathcal{X}_s, P(X_s)\}$, and the target task $T_t = \{\mathcal{Y}_t, \mathcal{F}_t(\cdot)\}$ with limited labelled training data. In this work, the proposed GLA-DA aims to improve the performance of target function $\mathcal{F}_t$ to accomplish the target task $T_t$ based on $\mathcal{D}_s$ and $T_s$while $\mathcal{Y}_s = \mathcal{Y}_t$.

\subsection{Pretraining}
In this work, the source encoder $M_s$ and the source classifier $C_s$ with labeled source data were trained in the pre-training process. The classification optimization goal is defined as,
\begin{equation}
      \min_{M_s,C_s}(\mathbf{X_s},{Y_s})\mathcal{L}_{C_s} \! = \! -\mathbb{E}_{(\mathbf{x_s},y)\sim(\mathbf{X_s}\!,\!{Y_s})}\!\sum_{k=0}^{K-1}\!\mbox{II}_{[y_s=k]} \mbox{log} C_s(\!M_s(\mathbf{x_s})\!)
    \label{eq:mcs}
\end{equation}
where $M_s$ and $C_s$ are the source encoder and source classifier respectively, $X_s$ and $Y_s$ are the source measurements and labels that have $K$ classes respectively.

\subsection{Assign Pseudo Labels}
\label{subsec:pseudo-labels}
Inspired by the agreement mechanism \cite{Li S: agreement mechanism} and injection mechanism\cite{Rawat: insertion mechanism}, we combined the advantages and characteristics of the two methods to obtain as many reliable pseudo labels as possible. In the semi-supervised scenario, scarce target labels are offered while in the unsupervised scenario, the labeled target data in the beginning are provided by the source model with those whose prediction possibility is greater than a threshold.

Specifically, we first used a Similarity-Based Classifier (SBC) which is a semi-supervised clustering model that captures the inner relationship among target samples to predict the labels of all target samples and thus get the labels of the unlabeled ones, where the LabelSpreading was adopted to leverage inner distribution information in our implementation. However, this similarity-based classifier could differ when applying GLA-DA to other domains of TS data. We denote these predicted labels as $y^{<SBC>}$. To refine the pseudo labels, we initialize target encoder $M_t$ as $M_s$ and then fine-tune it and the classifier on the labeled target data. The optimization goal of fine-tuning is defined in Eq. \ref{eq:fine-tune}. Next, we use the fine-tuned encoder and classifier, which are deep neural network (DNN) models, to predict the labels of the remaining unlabeled target data instances, denoted by $y^{<DNN>}$.

\begin{equation}
         \min_{M_t,C_s}(\mathcal{T}_L)\mathcal{L}_{ft} \! = \! -\mathbb{E}_{(\mathbf{x_t},y)\sim(\mathcal{T}_L)}\!\sum_{k=0}^{K-1}\!\mbox{II}_{[y_t=k]} \mbox{log} C_s(\!M_t(\mathbf{x_t})\!)
    \label{eq:fine-tune}
\end{equation}

To utilize target samples as many as possible, we injected an unlabeled target sample $i$ from $\mathcal{T}_U$ into $\mathcal{T}_{L}$ once Eq. \ref{eq:agree} is satisfied.

\begin{equation}
    y^{<SBC>}_i = y^{<DNN>}_i
    \label{eq:agree}
\end{equation}

With sufficient iterations, samples that are still unlabeled will be abandoned since $\mathcal{T}_L$ has been large enough and close to the whole target data set. 

\subsection{Global Feature Alignment}
\label{subsec:gfa}
By referring back to the process of evolution of adversarial deep transfer learning, there are two ways of adapting source and target domains: one way is to align one domain with the other, where the other domain is not adjusted; the other way is to make both domains be embedded into a third feature space. GLA-DA adopts the latter way. 

Specifically, first, we train a discriminator that tells which domain the sample is from. Then the target encoder is initialized with the parameters of the source encoder. Next, we train the target encoder and fine-tune the source encoder to deceive the discriminator. That's making the embedded data from the source and target obtain the same labels in the new domain.

Here, we assume labels of source and target domains and the shared space as 0, 1, and 0.5, respectively. The adversarial adaptation can be achieved by the optimization of the following three equations:

\begin{equation}
    \min_{M_s}\mathcal{L}_{M_s}(\mathbf{X_s},\mathbf{X_t},D) = -\mathbb{E}_{\mathbf{x_s} \sim \mathbf{X_s}}[\mbox{log}D(M_s(\mathbf{x_s}))]
    \label{eq9}
\end{equation}
\begin{equation}
    \min_{M_t}\mathcal{L}_{M_t}(\mathbf{X_t},\mathbf{X_t},D) = -\mathbb{E}_{\mathbf{x_t} \sim \mathbf{X_t}}[\mbox{log}D(M_t(\mathbf{x_t}))]
    \label{eq10}
\end{equation}
\begin{equation}
  \begin{split}
    \max_{D}\mathcal{L}_D(\mathbf{X_s},\mathbf{X_t},M_s,M_t) = \mathbb{E}_{\mathbf{x_s} \sim \mathbf{X_s}}[\mbox{log}(0.5- 
 \\ D(M_s(\mathbf{x_s})))]
    + \mathbb{E}_{\mathbf{x_t} \sim \mathbf{X_t}}[\mbox{log}(D(M_t(\mathbf{x_t}))-0.5)]
    \label{eq11}
  \end{split}  
\end{equation}  

where Eqs. \ref{eq9} and \ref{eq10} are the processes of training the discriminator, and Eq. \ref{eq11} is the process of training the target encoder and fine-tuning the source encoder. 

\subsection{Local Class Alignment}
\label{subsec:lca}
To achieve class-sensitive adaptation as shown in Fig. \ref{fig:distris} (b), we introduce a center-loss optimizer defined in Eq. \ref{eq:ctloss}, which calculates the square of the distance between the feature center point and every single data point. 
The center loss is proposed by Y Wen et al. \cite{Y Wen:Center loss}. It is featured by minimizing distances among data instances of the same class and maximizing those of different classes. Therefore, we adopted this center-loss optimization to avoid the class-disappearing situation shown in Fig. \ref{fig:distris} (a).

\begin{equation}
    \mathcal{L}_{ct} = \frac{1}{2}\sum_{i=1}^{batch\_size}||\mathbf{x_i}-\mathbf{c}_{y_i}||^{2}_{2}
    \label{eq:ctloss}
\end{equation}

Eq. \ref{eq:ctloss} optimizes parameters of source and target encoders to make data points of the same category closer to its center. Eq.\ref{eq:par} defines the gradient of $\mathcal{L}_{ct}$ and Eq.\ref{eq:del_cj} shows how to update the class difference $c_{y_i}$.

\begin{equation}
    \frac{\partial\mathcal{L}_{ct}}{\partial\mathbf{x_i}} = \mathbf{x_i}-\mathbf{c}_{y_i}
    \label{eq:par}
\end{equation}

\begin{equation}
    \Delta\mathbf{c_{j}} = \frac{\sum_{i=1}^{batch\_size}\mbox{II}(y_i=j)(\mathbf{c_j}-\mathbf{x_j})}{1+\sum_{i=1}^{batch\_size}\mbox{II}(y_i=j)}
    \label{eq:del_cj}
\end{equation}

\subsection{Shared Classifier}
After adversarial adaptation and center-loss optimization, the next step is to train a shared classifier $C_{sh}$ to classify the embedded data from source and target domains. The optimization goal is defined in Eq.\ref{eq:opti_sh}.

\begin{equation}
    \begin{split}
        \min_{C_{sh}}\mathcal{L}_{C_{sh}}\!(\mathbf{X_s},\!\mathbf{Y_s},\!\mathbf{X_t},\!{Y_t}) = -\mathbb{E}_{(\mathbf{x_s},y)\sim(\mathbf{X_s},\!{Y_s})}\!\sum_{k=0}^{K-1}\!\mbox{II}_{[k=y_s]} \\ \mbox{log} C_{sh}(M_s(\mathbf{x_s})) \! - \! \mathbb{E}_{(\mathbf{x_t},y)\!\sim(\mathbf{X_t},{Y_t})}\sum_{k=0}^{K-1}\mbox{II}_{[k=y_t]}\!\mbox{log} C_{sh}(\!M_t(\!\mathbf{x_t}\!)\!)
    \end{split}
    \label{eq:opti_sh}
\end{equation}

In summary, the final optimization goal of GLA-DA can be formulated as,

\begin{equation}
    \mathcal{L}_{GLA-DA} = \mathcal{L}_{C_s} + \mathcal{L}_{ft} + \mathcal{L}_{M_s} + \mathcal{L}_{M_t} + \mathcal{L}_{D} + \mathcal{L}_{ct} + \mathcal{L}_{C_{sh}}
    \label{eq:csda}
\end{equation}

The training process of GLA-DA is thus equivalent to the following objective:

\begin{equation}
    \min_{C_{sh}}\min_{M_s, M_t}\max_{D}\min_{M_t,C_s}\min_{M_s,C_s}\mathcal{L}_{GLA-DA}(\mathbf{X_s},\mathbf{X_t},{Y_s},Y_t,D,M_s,M_t)
\end{equation}

\section{Experiments}
\label{sec: exp}
\subsection{Dataset}
We evaluated the performance of the proposed GLA-DA model on 4 datasets across 2 real-world application scenarios, including human activity recognition (UCIHAR, HHAR, and WISDM) and sleep stage classification (EEG from sleep-EDF). Here EEG is univariate for verifying our proposed GLA-DA is also suitable for univariate TS data and the other are multivariate. Codes are available at https://github.com/TuGang-Git/GLA-DA-for-Multivariate-Time-Series.


For the three datasets for human activity recognition (HAR), UCIHAR collects 30 persons' data from sensors in smartphones \cite{UCIHAR}. HHAR (Heterogeneous HAR) collects data from smartphones and smartwatches \cite{HHAR}. Samples collected by the 2 devices are heterogeneous and we only adopted data from smartphones to conduct experiments, as previous research \cite{Ragab: ADATIME} did. WISDM (Wireless Sensor Data Mining) collected data from accelerometer sensors \cite{WISDM}. Volunteers in data collection of the three datasets were asked to perform 6 actions: walking, walking downstairs, walking upstairs, sitting, standing, and lying. The last dataset we used is for sleep stage recognition, named sleep-EDF. It had 5 stages: Wake, Non-Rapid Eye Movement (N1, N2, N3), and Rapid Eye Movement, which can be determined by electroencephalography (EEG) signals \cite{EEG}. We chose the single EEG channel in our experiments following \cite{Eldele: ADLSSC}. The input channels of the 4 datasets are 9,3,3 and 1 respectively.

\subsection{Setups}
\subsubsection{Dataset Preprocessing}
The dataset was divided into a ratio of $7:3$ for training and testing, respectively. In the semi-supervised scenario, since the datasets are imbalanced, we kept labels of each class at the same ratio randomly. Note that, for simplicity, we will use 'Source $\rightarrow$ Target' to indicate the adaptation from the source domain to the target domain. 

\subsubsection{Software and Hardware Environment} We conducted our experiments on a computer having a CPU of 13th Gen Intel(R) Core(TM) i7-13700F and a GPU of RTX 4060, using Python 3.11 with Pytorch 2.0.1 as the programming framework and CUDA 11.8 for faster GPU computation. 

\subsubsection{Model Architecture}
\begin{figure}[!t]
    \centering
    \includegraphics[width=1.0\linewidth]{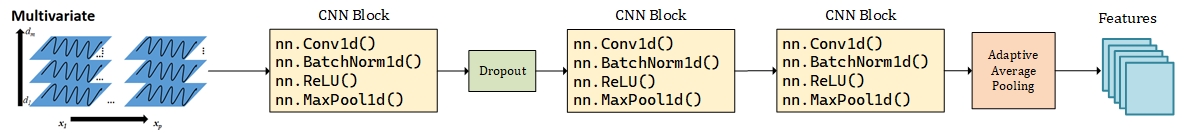}
    \caption{Architecture of 1D-CNN encoder used in our proposed GLA-DA.}
    \label{fig:cnn}
\end{figure}

We chose a 1D convolutional neural network (1D-CNN) as the backbone encoder, an (NN) of single fully connected (FC) layer as the classifier followed by \cite{Ragab: ADATIME} and an NN of 3 FC layers as the domain discriminator. As shown in Fig.\ref{fig:cnn}, the 1D-CNN we adopted has three convolution blocks, with a dropout layer followed by the first block and an adaptative average pooling layer at the end. Each block consists of a 1D-Convolution layer, a batchnorm layer, an activate function (non-linear ReLU), and a max pooling layer subsequently. Related hyper-parameters will be discussed in the next part.

\subsubsection{Hyper-parameters}
We have several hyper-parameters in the training process and models. In training, we set the epoch of pre-training, assigning pseudo labels and adaptation as 40, 3, and 50 respectively, the threshold mentioned in Sec.\ref{sec: method} for obtaining initial labeled target instances as 0.7, learning rate for source encoder, target encoder, discriminator, classifier and center-loss-optimizer as 0.0001, 0.00005, 0.001, 0.001 and 0.001 respectively. We adopted Adam as the optimizer and set the beta pair as (0.5, 0.9). As for the models, we set a discriminator consisting of an input and hidden layer of 128 neurons and an output layer of 1 neuron and a classifier consisting of an input layer of 128 neurons and an output layer of 5 neurons for EEG and 6 for the left. For the 1D-CNN encoder, we set the kernel, stride, and padding pair as (2,2,1) in the max-pooling layer, dropout rate as 0.2 for EEG and 0.5 for the left, (kernel, stride, padding) pair as (26,5,13) for EEG and (5,1,2) for the left in the first convolution layer and (8,1,4) in subsequent convolution layers and (2,2,1) for the max-pooling layer. The intermediate channels are 32 for EEG and 64 for the left \cite{Eldele: CoMix}.

\subsection{Baselines}
We adopted Source Only and 4 SOTA DA methods covering contrastive and adversarial learning on TS data:

\begin{itemize}
    \item[·] \textbf{Source Only}: Applying source model on target data directly;
    \item[·] \textbf{CoDATS}: Convolutional deep Domain Adaptation model for TS data\cite{Wilson: CoDATS}; 
    \item[·] \textbf{AdvSKM}: Adversarial Spectral Kernel Matching \cite{Liu: AdvSKM};
    \item[·] \textbf{SASA}:TS DA via sparse associative structure alignment \cite{Cai: SASA};
    \item[·] \textbf{CoTMix}: Contrastive Domain Adaptation for TS via Temporal Mixup \cite{Eldele: CoMix}
    
\end{itemize}

\subsection{Results and Comparison}
Due to the imbalance of the datasets used in experiments, we evaluated our GLA-DA model and compared it to baselines based on the macro-F1 (MF1) score. The MF1 score is the mean of the F1-scores for each class defined in \ref{eq:f1}:

\begin{equation}
      \text{F1-score} = 2 \times \frac{\text{Precision} \times \text{Recall}}{\text{Precision}+\text{Recall}}
    \label{eq:f1}
\end{equation}

where Precision is the ratio of true positive predictions to the total predicted positives and Recall is the ratio of true positive predictions to the total actual positives. With f1-score of each class, we can get MF1-score: $\frac{1}{N}\sum_{i=1}^N\text{F1-score}_i$ where N is the number of classes. We denote 'GLA-DA' as the unsupervised implementation of our proposed method and 'GLA-DA (1\% labeled) as the semi-supervised one. Results across 20 scenarios are shown in Table \ref{tab:results}.

\begin{table}[!h]
\centering
\caption{Experimental results on MF1-score across 20 DA scenarios}
\begin{tabular}{c|c|ccccc|cc}
\hline
Dataset                & Scenario & \begin{tabular}[c]{@{}c@{}}Source \\ Only\end{tabular} & CoDATS         & AdvSKM & SASA  & CoTMix         & \begin{tabular}[c]{@{}c@{}}GLA-DA\\ (1\% labeled)\end{tabular} & GLA-DA         \\ \hline
\multirow{5}{*}{HAR}   & 7 → 13   & 81.71       & 93.33          & 92.25  & 91.13 & \textbf{98.96} & 97.88                                                          & 95.68          \\
                       & 9 → 18   & 58.34       & 91.50           & 67.86  & 76.51 & 89.42          & 96.39                                                          & \textbf{96.83} \\
                       & 22 → 4   & 50.60        & 86.72          & 89.28  & 83.34 & \textbf{96.93} & 96.87                                                          & 95.82          \\
                       & 4 → 9    & 39.40        & 57.32          & 53.55  & 66.80  & 58.41          & \textbf{90.28}                                                 & 88.03          \\
                       & 12 → 10  & 49.82       & 45.75          & 43.58  & 62.06 & 45.56          & 88.15                                                          & \textbf{92.37} \\ \hline
\multirow{5}{*}{HHAR}  & 1 → 6    & 80.12       & 91.97          & 62.36  & 89.95 & 89.89          & \textbf{97.02}                                                 & 96.61          \\
                       & 4 → 5    & 76.47       & 79.33          & 80.21  & 95.81 & 97.57          & 98.67                                                          & \textbf{99.03} \\
                       & 7 → 4    & 55.13       & 95.18          & 83.70  & 90.98 & 91.62          & \textbf{98.13}                                                 & 90.20           \\
                       & 0 → 3    & 42.96       & 81.00          & 61.09  & 70.07 & 57.84          & \textbf{98.89}                                                 & 96.30           \\
                       & 1 → 5    & 92.85       & 97.56          & 95.77  & 96.32 & 97.94          & 98.28                                                          & \textbf{98.64} \\ \hline
\multirow{5}{*}{WISDM} & 7 → 17   & 60.03       & 81.56          & 52.31  & 60.37 & 72.00          & \textbf{96.35}                                                 & 93.36          \\
                       & 20 → 30  & 53.18       & 68.78          & 66.54  & 61.76 & 76.02          & \textbf{94.43}                                                 & 92.80           \\
                       & 6 → 19   & 46.80        & 65.95          & 49.5   & 38.47 & 47.44          & \textbf{99.15}                                                 & 92.94          \\
                       & 18 → 23  & 56.70        & 61.74          & 55.57  & 42.41 & 80.61          & 90.57                                                          & \textbf{93.83} \\
                       & 32 → 30  & 31.21       & 76.63          & 59.38  & 59.66 & 66.64          & 95.2                                                           & \textbf{98.91} \\ \hline
\multirow{5}{*}{EEG}   & 16 → 1   & 45.36       & 64.10          & 61.66  & 58.92 & 56.38          & 63.12                                                          & \textbf{65.86} \\
                       & 0 → 11   & 53.84       & 37.96          & 43.23  & 39.99 & 45.16          & \textbf{65.83}                                                 & 62.47          \\
                       & 13 → 1   & 45.39       & \textbf{76.90} & 53.35  & 66.05 & 54.01          & 53.47                                                          & 58.07          \\
                       & 12 → 4   & 46.57       & 53.27          & 46.58  & 49.41 & 45.86          & 72.61                                                          & \textbf{75.01} \\
                       & 19 → 9   & 66.55       & 61.01          & 67.67  & 68.64 & 67.11          & \textbf{83.48}                                                 & 71.83          \\ \hline
\end{tabular}
\label{tab:results}
\end{table}

From the results, we can see that  Source Only had the worst performance in most cases. This can be explained by the difference in data distribution that can lead to domain shift problems. Our proposed GLA-DA outperforms the SOTA methods in most scenarios, including its unsupervised and semi-supervised implementation, where the unsupervised one stands the best for 8 scenarios and the semi-supervised one for 9 scenarios. Only in three scenarios did GLA-DA not perform better than the baselines but maintained a gap of no more than 4\% except in 13 → 1 of the EEG dataset. Besides, all baselines show good performance and some of them are very close to the best like scenario 7 → 4 of HHAR (CoDATS) and 16 → 1 of EEG (CoDATS), this can be attributed to the attributes of 1D-CNN, which excels at capturing temporal dependencies. The results indicate the superiority of GLA-DA on multivariate time series data and prove its ability to generalization to the univariate TS data. 

To evaluate the performance of our proposed AM, we also calculated the MF1-score of generated pseudo labels, and results are provided in Table \ref{tab:pseudolabels}.

\begin{table}[!t]
\caption{Results of MF1-scores of pseudo labels. MF1-score (1\% labeled) indicates the semi-supervised scenario.}
\resizebox{\textwidth}{!}{
\begin{tabular}{c|ccc|ccc|ccc|ccclllllll}
\cline{1-13}
 & \multicolumn{3}{c|}{HAR} & \multicolumn{3}{c|}{HHAR} & \multicolumn{3}{c|}{WISDM} & \multicolumn{3}{c}{EEG} &  &  &  &  &  &  &  \\ \cline{1-13}
Scenario & \multicolumn{1}{c|}{7 → 13} & \multicolumn{1}{c|}{22 → 4} & 12 → 10 & \multicolumn{1}{c|}{1 → 6} & \multicolumn{1}{c|}{7 → 4} & 1 → 5 & \multicolumn{1}{c|}{7 → 17} & \multicolumn{1}{c|}{20 → 30} & 18 → 23 & \multicolumn{1}{c|}{0 → 11} & \multicolumn{1}{c|}{13 → 1} & 19 → 9 &  &  &  &  &  &  &  \\ \cline{1-13}
\begin{tabular}[c]{@{}c@{}}MF1-score\\ (1\% labeled)\end{tabular} & \multicolumn{1}{c|}{\textbf{82.49}} & \multicolumn{1}{c|}{\textbf{93.49}} & 46.94 & \multicolumn{1}{c|}{\textbf{92.79}} & \multicolumn{1}{c|}{\textbf{91.80}} & 91.67 & \multicolumn{1}{c|}{\textbf{70.81}} & \multicolumn{1}{c|}{\textbf{58.18}} & 58.05 & \multicolumn{1}{c|}{\textbf{57.01}} & \multicolumn{1}{c|}{\textbf{53.87}} & \textbf{63.53} &  &  &  &  &  &  &  \\ \cline{1-13}
MF1-score & \multicolumn{1}{c|}{77.33} & \multicolumn{1}{c|}{72.16} & \textbf{54.26} & \multicolumn{1}{c|}{89.48} & \multicolumn{1}{c|}{70.79} & \textbf{97.03} & \multicolumn{1}{c|}{44.73} & \multicolumn{1}{c|}{54.02} & \textbf{69.53} & \multicolumn{1}{c|}{55.72} & \multicolumn{1}{c|}{40.35} & 52.38 &  &  &  &  &  &  &  \\ \cline{1-13}
\end{tabular}
}
\label{tab:pseudolabels}
\end{table}

It is worth noting that in some cases the unsupervised implementation exceeds the semi-supervised one in terms of the Mf1-score metric for the final classification task and assigning pseudo labels as is mentioned in Sec \ref{sec: method}, in an unsupervised scenario, the initial labeled target instances are obtained by the source model while in the semi-supervised scenario, they are offered in the beginning. As Table \ref{tab:results} shows, although Source Only suffers from domain shift, it can still generate some correct labels. This phenomenon may cause the unsupervised scenario to have more target instances that are correct and helpful for later training than those in the semi-supervised scenario. This can be supported by data from 12 → 10 of HAR and 1 → 5 of HHAR in Table \ref{tab:pseudolabels}.

\begin{figure}[!t]
    \centering
    \includegraphics[width=1.0\linewidth]{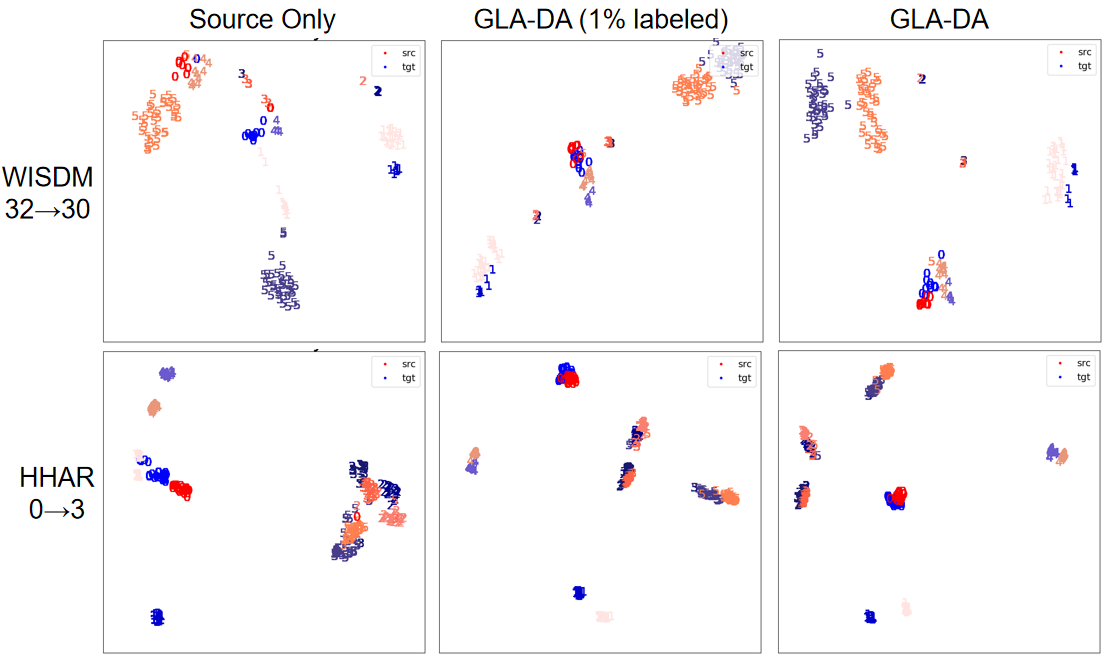}
    \caption{Effect of local class alignment compared to Source Only. Red and its different color depth indicates the source domain and blue and its different color depth indicates the target domain.  }
    \label{fig:distribution}
\end{figure}

To verify the effect of local class alignment, we visualized the embedded features distribution after adaptation by Source Only and two implementations of GLA-DA with t-SNE. Results in scenario 32 → 30 of WISDM and 0 → 3 of HHAR are provided in Fig \ref{fig:distribution}. As we can see, only the data instances from the same classes are aligned close to each other after the adaptation by the proposed GLA-DA, which utilized center loss to keep differences among inner classes. This visualization results show that after LCA, data instances have a clear decision boundary within each class, which can be helpful for the training procedure of the shared classifier and avoid potential overfitting.

\subsection{Ablation Experiments}
To ensure the importance of the key step, LCA, we also conducted experiments without (w/o) this step. Results are shown in Table \ref{tab:ablation}. We can see that in all presented scenarios, the original semi-supervised implementation achieves better performance than that without LCA and the biggest gap between them comes to  42\% (12 → 10 of HAR). This strongly proves the necessity of LCA, which can pull data instances of the same class closer and those of different classes farther. 

\begin{table}[!t]
\caption{Results of ablation experiments}
\label{tab:ablation}
\resizebox{\textwidth}{!}{
\begin{tabular}{c|ccc|ccc|ccc|ccclllllll}
\cline{1-13}
 & \multicolumn{3}{c|}{HAR} & \multicolumn{3}{c|}{HHAR} & \multicolumn{3}{c|}{WISDM} & \multicolumn{3}{c}{EEG} &  &  &  &  &  &  &  \\ \cline{1-13}
Scenario & \multicolumn{1}{c|}{20 → 28} & \multicolumn{1}{c|}{4 → 9} & 12 → 10 & \multicolumn{1}{c|}{4 → 5} & \multicolumn{1}{c|}{0 → 3} & 1 → 6 & \multicolumn{1}{c|}{7 → 17} & \multicolumn{1}{c|}{18 → 23} & 32 → 30 & \multicolumn{1}{c|}{16 → 1} & \multicolumn{1}{c|}{15 → 4} & 19 → 9 &  &  &  &  &  &  &  \\ \cline{1-13}
\begin{tabular}[c]{@{}c@{}}GLA-DA\\ (1\% labeled)\end{tabular} & \multicolumn{1}{c|}{\textbf{94.19}} & \multicolumn{1}{c|}{\textbf{90.28}} & \textbf{88.15} & \multicolumn{1}{c|}{\textbf{98.67}} & \multicolumn{1}{c|}{\textbf{98.89}} & \textbf{97.02} & \multicolumn{1}{c|}{\textbf{96.35}} & \multicolumn{1}{c|}{\textbf{90.57}} & \textbf{95.20} & \multicolumn{1}{c|}{\textbf{63.12}} & \multicolumn{1}{c|}{\textbf{75.24}} & \textbf{83.48} &  &  &  &  &  &  &  \\ \cline{1-13}
\begin{tabular}[c]{@{}c@{}}GLA-DA\\ (1\% labeled)\\ w/o LCA\end{tabular} & \multicolumn{1}{c|}{73.92} & \multicolumn{1}{c|}{83.31} & 62.07 & \multicolumn{1}{c|}{94.89} & \multicolumn{1}{c|}{76.08} & 94.56 & \multicolumn{1}{c|}{78.27} & \multicolumn{1}{c|}{71.97} & 79.94 & \multicolumn{1}{c|}{57.68} & \multicolumn{1}{c|}{72.15} & 73.00 &  &  &  &  &  &  &  \\ \cline{1-13}
\end{tabular}
}
\end{table}

\section{Conclusion}
\label{sec:conc}
In this work, we proposed a Global-Local Alignment Domain Adaptation (GLA-DA) method for multivariate time series data with 1D-CNN as the backbone encoder to capture temporal dependencies, which implements global feature alignment by projecting data from both domains to an intermediate feature space in an adversarial manner and addresses local class alignment to keep the differences of distribution among different classes by a center-loss optimizer. To enhance GLA-DA's performance, we also proposed a novel agreement mechanism to assign pseudo labels by the consistency between a similarity-based model and a deep learning-based model. We implemented GLA-DA in an unsupervised and semi-supervised scenario and conducted various experiments across 20 scenarios in real-world applications of human activity recognition and sleep stage classification with macro f1-score as metric to cope with data-imbalance issue and showed the superiority of GLA-DA over baselines. Besides, we also showed the visualization results of the data distribution to verify the effect and conducted ablation experiments to prove the necessity of local class alignment. Experimental results also show that our method can be expanded to be applied to univariate time series data. Future work could further discuss the other methods for assigning pseudo labels and aligning data distributions globally and locally.


%
%
%
%

\end{document}